\documentclass[letterpaper, 10 pt, journal, twoside]{./IEEEtran}

\usepackage{graphicx}
\usepackage{color}
\usepackage{cite}
\usepackage[hidelinks,bookmarks=false]{hyperref}
\usepackage{multirow}
\usepackage{enumerate}
\usepackage{bm}
\usepackage{booktabs}
\usepackage{blindtext}
\usepackage[T1]{fontenc}
\usepackage{mathptmx} 
\DeclareMathAlphabet{\mathcal}{OMS}{cmsy}{m}{n}
\usepackage{times} 
\usepackage{amsmath} 
\usepackage{amssymb}  
\usepackage[ruled,linesnumbered]{algorithm2e}
\usepackage{url}
\usepackage{bbm}
\usepackage{booktabs}
\usepackage{upgreek}
\usepackage{mathtools,xparse}
\usepackage[utf8]{inputenc}
\usepackage{subfigure}

\ifCLASSINFOpdf
\else
\fi
\DeclarePairedDelimiter{\norm}{\lVert}{\rVert}
\NewDocumentCommand{\normL}{ s O{} m }{%
  \IfBooleanTF{#1}{\norm*{#3}}{\norm[#2]{#3}}_{L_2($\Omega$)}%
}

\usepackage{stackengine}
\usepackage{scalerel}

\IEEEaftertitletext{\vspace{-1.0\baselineskip}}

\begin{document}

\makeatletter
\newcommand{\rmnum}[1]{\romannumeral #1}
\newcommand{\Rmnum}[1]{\expandafter\@slowromancap\romannumeral #1@}
\makeatother

\title{Probabilistic End-to-End Vehicle Navigation in Complex Dynamic Environments with\\ Multimodal Sensor Fusion}

\author{Peide~Cai,
        ~Sukai~Wang,
        ~Yuxiang~Sun,
        ~Ming~Liu,~\IEEEmembership{Senior Member,~IEEE}
\thanks{All authors are with The Hong Kong University of Science and Technology, Hong Kong SAR, China (email: pcaiaa@connect.ust.hk; swangcy@connect.ust.hk; sun.yuxiang@outlook.com; eelium@ust.hk).}%
}


\maketitle

\begin{abstract}
All-day and all-weather navigation is a critical capability for autonomous driving, which requires proper reaction to varied environmental conditions and complex agent behaviors. Recently, with the rise of deep learning, end-to-end control for autonomous vehicles has been well studied. However, most works are solely based on visual information, which can be degraded by challenging illumination conditions such as dim light or total darkness. In addition, they usually generate and apply deterministic control commands without considering the uncertainties in the future. In this paper, based on imitation learning, we propose a probabilistic driving model with multi-perception capability utilizing the information from the camera, lidar and radar. We further evaluate its driving performance online on our new driving benchmark, which includes various environmental conditions (e.g., urban and rural areas, traffic densities, weather and times of the day) and dynamic obstacles (e.g., vehicles, pedestrians, motorcyclists and bicyclists). The results suggest that our proposed model outperforms baselines and achieves excellent generalization performance in unseen environments with heavy traffic and extreme weather.
\end{abstract}

\begin{IEEEkeywords}
Automation technologies for smart cities, autonomous vehicle navigation, multi-modal perception, sensorimotor learning, motion planning and control.
\end{IEEEkeywords}

%
\IEEEpeerreviewmaketitle

\section{Introduction}
\IEEEPARstart{I}{n the} field of autonomous driving, traditional navigation methods are commonly implemented with modular pipelines\cite{leonard2008perception, dickmanns2002development}, which split the navigation task into individual sub-problems, such as perception, planning and control. These modules often rely on a multitude of engineering components to produce reliable environmental representations, robust decisions and safe control actions. However, since the separate modules rely on each other, the system can lead to an accumulation of errors. Therefore, each component requires careful and time-consuming hand engineering. 

\begin{figure}[thpb]
    \centering
    \includegraphics[width = \columnwidth]{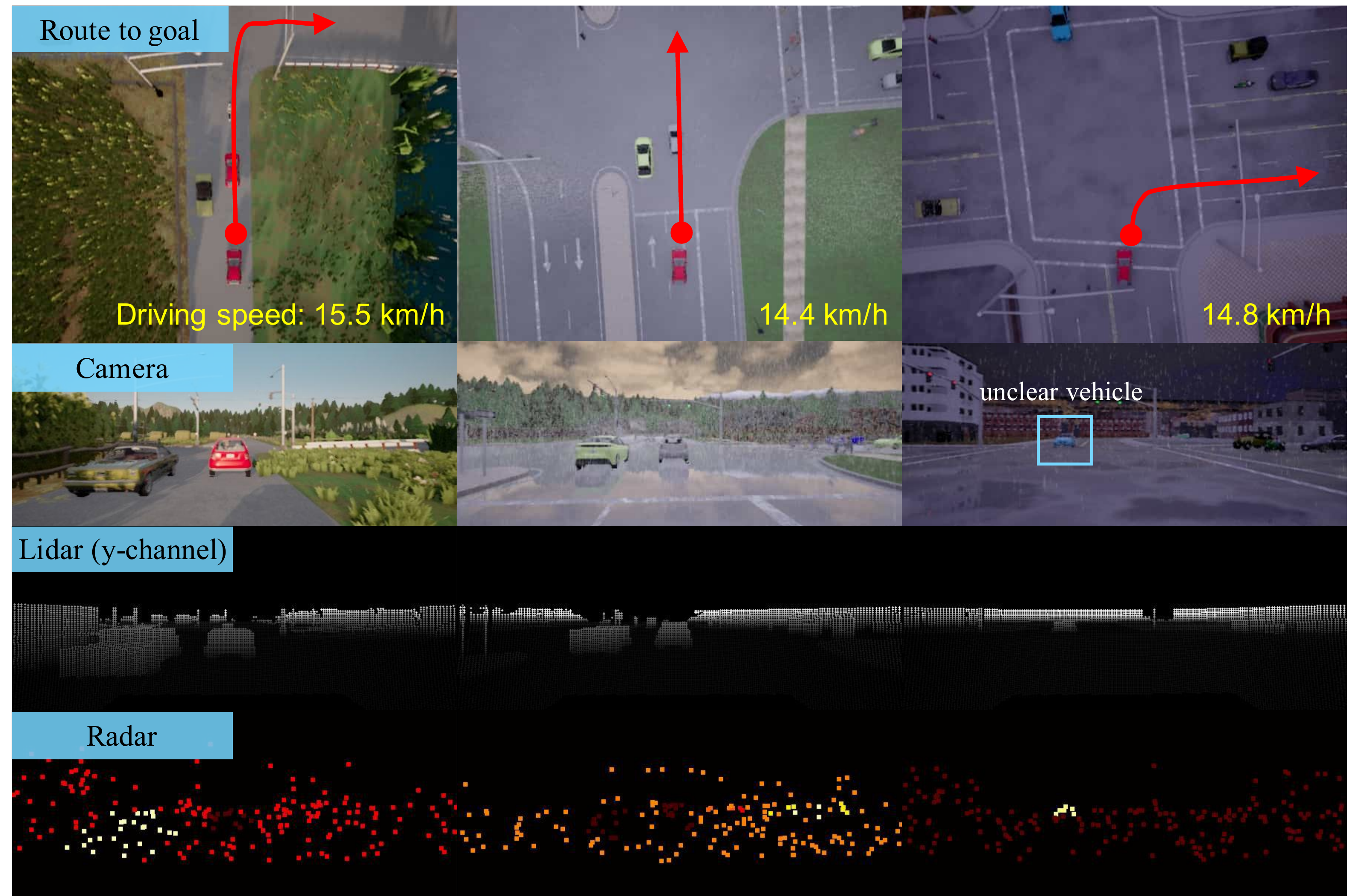}
    \caption{Snapshots of different driving scenarios (left to right: \textit{ClearDay}, \textit{RainySunset} and \textit{DrizzleNight}) with global route directions and sensor data information. For visualization, we project the lidar data (y-channel, i.e., the height information) and radar data (relative speed to the ego-vehicle) to the image plane. Brighter points mean larger values. It can be seen that the information characteristic from lidar and radar is more consistent than from the camera in different environmental conditions.}
    \label{scenarios}
    \vspace{-0.3cm}
\end{figure}

In recent years, with the unprecedented success of deep learning, an alternative method called end-to-end control\cite{bojarski2016end, gao2017intention, codevilla2018end, xu2017end, codevilla2019exploring, tai2019visual , amini2019variational, hecker2018end, karkus2019dan, muller2018driving} has arisen. This paradigm mimics the human brain and maps the raw sensory input (e.g., RGB images) to control output (e.g., steering angle) in an end-to-end fashion. In addition, it substitutes laborious hand engineering by learning control policies directly on data from human drivers with deep networks, where explicit programming or modeling of each possible scenario is not needed. Moreover, it can adapt to complex noise characteristics of different environments during training, which cannot be captured well by analytical methods.

While end-to-end driving has been considerably fruitful, there exist three critical deficiencies in the prior works.

1) The visual information is stressed too much. Most works depend solely on cameras for scene understanding and decision making\cite{bojarski2016end, gao2017intention, codevilla2018end, xu2017end, codevilla2019exploring, tai2019visual , amini2019variational, hecker2018end, karkus2019dan, muller2018driving, cai2019vision, cai2020vtgnet}. However, although cameras are versatile and cheap, they have difficulty capturing fine-grained 3-D information. In addition, perception relying on cameras is prone to be affected by challenging illumination and weather conditions, such as the \textit{DrizzleNight} case shown in Fig. \ref{scenarios}. Because of dim light and rain drops in this scene, the blue car far ahead left can be difficult to recognize. In such scenarios, vision-based driving systems can be dangerous. However, the blue car is quite distinguishable by observing the speed distribution from the radar data. 

2) The probabilistic nature of executable actions is not well explored. Most works output deterministic commands to the vehicle \cite{bansal2019chauffeurnet, pokle2019deep}; however, non-determinism is a key aspect of controlling, which is useful in many safety-critical tasks such as collision checking and risk-aware motion planning \cite{huang2019uncertainty}. A more reasonable approach, therefore, should be predicting a motion distribution indicating \textit{what could do} rather than \textit{what to do} for the driving platform.

3) The prior end-to-end methods are not evaluated sufficiently in terms of the \textit{navigation} task. Most works are evaluated by first collecting a driving dataset with ground-truth annotations (e.g., expert control actions) and then measuring the average prediction error \textit{offline} on the test set\cite{huang2019uncertainty, amini2019variational, xu2017end, hecker2018end, cai2019vision, cai2020vtgnet}. However, different from the computer vision tasks such as object detection, the priority of driving should be safety and robustness rather than accuracy. As indicated in \cite{codevilla2018offline}, the offline prediction error cannot well reflect the actual driving quality. Therefore, \textit{online} evaluation is more reasonable and should be given more attention. One critical concern for online evaluation is the environmental complexity, yet prior related works either test their methods in static maps \cite{pokle2019deep, pfeiffer2017from, karkus2019dan, muller2018driving, Cai2020HighSpeedAD}, or scenarios with low-level complexity \cite{codevilla2019exploring, tai2019visual,bansal2019chauffeurnet,gao2017intention, codevilla2018end, bojarski2016end}.

The aforementioned limitations motivate our exploration to enhance the perception capability for end-to-end driving systems. To this end, we propose a mixed sensor setup combining a camera, lidar and radar. The multimodal information is processed by uniform alignment and projection onto the image plane. Then, ResNet\cite{he2016resnet} is used for feature extraction. Based on this setup, we introduce a probabilistic motion planning (PMP) network to learn a deep probabilistic driving policy from expert provided data, which outputs both a distribution of future motion based on the Gaussian mixture model (GMM) \cite{wiest2012probabilistic, amini2019variational, huang2019uncertainty}, and a deterministic control action. Finally, we evaluate the driving performance of our model online on a new benchmark with extensive experiments. The main contributions of this letter are summarized as follows.

\begin{itemize}
    \item An end-to-end navigation method with multimodal sensor fusion and probabilistic motion planning, named PMP-net, for improving perception capability and considering uncertainties in the future.
    \item A new online benchmark, named \textit{DeepTest}, to perform analysis of driving systems in high-fidelity simulated environments with varied maps, weather, lighting conditions and traffic densities.
    \item Extensive evaluation and human-level driving performance of the proposed PMP-net, presented in unseen urban and rural areas with extreme weather and heavy traffic.
\end{itemize}

\section{Related Work}
End-to-end control is designed with deep networks to directly learn a mapping from raw sensory data to control outputs. The pioneer ALVINN system \cite{pomerleau1989alvinn} developed in 1989 uses a multilayer perceptron to learn the directions a vehicle should steer. With the recent advancement of deep learning, end-to-end control techniques have experienced tremendous success. For example, using more powerful modern convolutional neural networks (CNNs) and higher computational power, Bojarski \textit{et al.}\cite{bojarski2016end} demonstrate impressive performance in simple real-world driving scenarios such as on flat or barrier-free roads. Xu \textit{et al.}\cite{xu2017end} develop an end-to-end architecture to predict future vehicle egomotion from a large-scale video dataset. However, these works only realize a lane-following task and goal-directed navigation is not studied. 

To enable goal-directed autonomous driving, Codevilla \textit{et al.} \cite{codevilla2018end} propose a conditional imitation learning pipeline. In this work, the vehicle is able to take a specific turn at intersections based on high-level navigational commands such as \textit{turn left} and \textit{turn right}. Follow-up works include \cite{codevilla2019exploring, muller2018driving, cai2019vision} and \cite{cai2020vtgnet}. Another trend of adding guidance to the control policy is using global route, which is a richer representation of the intended moving directions than turning commands. For example, Gao \textit{et al.} \cite{gao2017intention} render routes on 2D floor maps and call them \textit{intentions}. Then, a neural-network motion controller maps \textit{intentions} and camera images directly to robot actions. Pokle \textit{et al.} \cite{pokle2019deep} follow this idea and implement a deep local trajectory planner and a velocity controller to compute motion commands based on the path generated by a global planner. However, these two works only focus on indoor robot navigation. For outdoor driving applications, Cai \textit{et al.}\cite{Cai2020HighSpeedAD} realize high-speed autonomous drifting in racing scenarios guided by route information with deep reinforcement learning. However, the control policy is only evaluated in static maps. Hecker \textit{et al.} \cite{hecker2018end} propose to learn a control policy with GPS-based route planners and surround-view cameras. However, as with many other works \cite{huang2019uncertainty, amini2019variational, xu2017end, cai2019vision}, this work is only evaluated offline by analysing the average predicting error, providing unclear information of the actual driving quality.

Inspired by the route-guided navigation methods mentioned above, we use a global planner to compute paths towards destinations in outdoor driving areas. For the low-level reactive control, we implement an end-to-end network translating the global route into driving actions (steering, throttle and brake). Based on this architecture, point-to-point autonomous driving can be realized. The network is trained with imitation learning and can adapt to varied environments to drive appropriately (e.g., slow down at intersections) and safely (e.g., slow down for a car, and urgently stop for jaywalkers). Similar to \cite{ gao2017intention} and \cite{pokle2019deep}, we assume that the localization information is available during system operation. However, different to \cite{ gao2017intention} and \cite{pokle2019deep}, our work focuses on complicated outdoor driving scenarios, and combines multimodal sensors complementing each other to generate unified perception results. 

In addition, our approach relates to the work of probabilistic driving models. To improve the capability of handling long-term plans with imitation learning, Amini \textit{et al.} \cite{amini2019variational} propose a variational network to predict a full distribution over possible steering commands. Similarly, Huang \textit{et al.} \cite{huang2019uncertainty} propose to use GMM to predict a distribution of future vehicle trajectories. These works explicitly consider uncertainties of future motions on logged data with \textbf{offline} metrics. By contrast, we evaluate our probabilistic driving model \textbf{online} with varied environmental conditions (e.g., rainy night with heavy traffics), which has not been studied in this context before.

\begin{figure*}[t]
    \centering
    \includegraphics[width = 2\columnwidth]{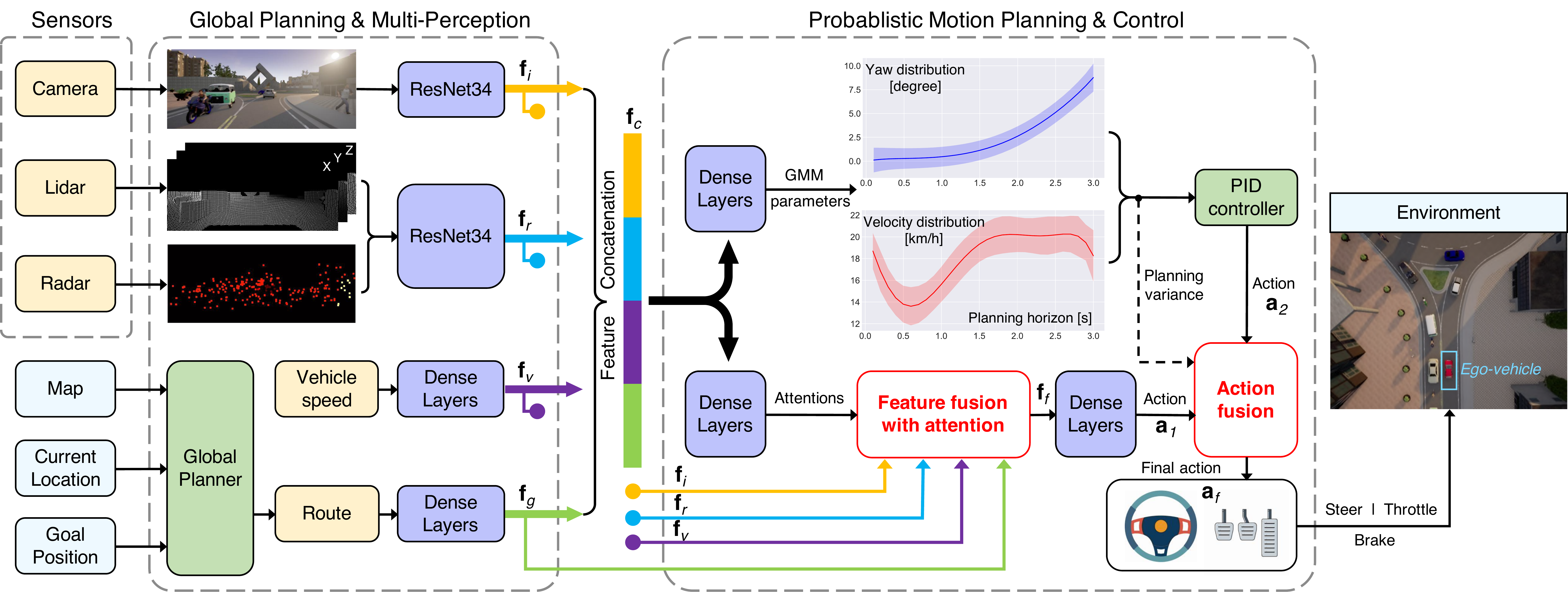}
    \caption{The architecture of our probabilistic motion planning network (PMP-net). It receives the multimodal sensory input and plans a motion distribution for 3 seconds in the future, based on which a PID controller is designed to generate a control action ${a_2}$. In addition, PMP-net generates another action ${a_1}$ in an end-to-end fashion. Then the variance of the planned motion distribution is used to fuse the dual actions for controlling the vehicle.}
    \label{network}
    \vspace{-0.3cm}
\end{figure*}

\section{Methodology}
\subsection{Formulation}
We formulate the problem of autonomous vehicle navigation as a goal-directed motion planning task to be solved by an end-to-end network architecture with imitation learning. The goal is to control the vehicle to drive safely and robustly in complex outdoor areas to achieve point-to-point navigation, like a human driver. To this end, we design a probabilistic driving model using multimodal perceptions from the camera, lidar and radar. In addition, we choose the latest CARLA simulation (0.9.7) \cite{dosovitskiy2017carla} to train and evaluate the system\footnote{Different from the older versions of CARLA (0.8.x) used in \cite{codevilla2018end, codevilla2019exploring} and \cite{tai2019visual}, which contain only two urban maps, the latest CARLA environment provides seven maps covering both urban and rural areas, with more available sensors, improved physical dynamics and more realistic illuminations. \url{http://carla.org/2019/12/11/release-0.9.7/}}. The entire pipeline of our PMP-net is shown in Fig. \ref{network}.

\subsection{Dataset Collection}
\label{dataset_collection}
To make the model successfully learn the knowledge of goal-directed reactive control in the context of outdoor driving, we collect a large-scale dataset with a global planner and an expert demonstrator in CARLA. At the beginning of each driving episode, the ego-vehicle is spawned at a random position $\bm{p}$. Then a collision-free coarse route (ranging from 350 m to 1500 m) from $\bm{p}$ to a destination $\bm{d}$ is provided by a global planner. The vehicle then follows this route at a speed of around 40 km/h while reacting to local environments to avoid collisions, such as slowing down for a forward-facing car that is moving slowly. Additionally, the vehicle reasonably slows the speed down to 15 km/h at intersections to ensure safety. In the process of data collection, we record the vehicle velocities, yaw angles, RGB images, lidar/radar data and expert driving actions (i.e., steering, throttle and brake) at 10 Hz. Moreover, in order to increase the complexity of our dataset, we focus on the following two aspects:

\subsubsection{Complexity of Environments} a) The datasets from prior works\cite{tai2019visual, codevilla2018end, codevilla2019exploring} are generated only in one map with two lanes and 90-degree turns (\textit{Town01} in Fig. \ref{dataset}). By contrast, we use five urban maps for data collection, which consist of different types of intersections and even roundabouts, and multiple lanes on roads; b) We set nine combinations of weather (\textit{clear}, \textit{drizzle} and \textit{rainy}) and illumination (\textit{daytime}, \textit{sunset} and \textit{night}). Heavier rain leads to more puddles on roads, and thus brings a greater reflection effect for visual perception.

\subsubsection{Complexity of Road Agents} a) We set pedestrians with different appearances (children and adults) randomly running or walking along the sidewalks and crosswalks. They occasionally disobey traffic rules and cross the road abruptly without previous notice, which increases the safety burden for autonomous driving; b) We set different types of vehicles (e.g., cars, trucks, vans, jeeps, buses, motorcyclists and bicyclists) with multiple appearances navigating around the cities at varied speeds. Based on a) and b), we apply four levels of traffic density for data collection: \textit{empty}, \textit{few}, \textit{regular} and \textit{dense}. Note that these road agents are controlled by the AI engine from CARLA to construct realistic city scenarios.

\begin{figure*}[t]
    \centering
    \setlength{\abovecaptionskip}{-1pt}
    \includegraphics[width = 2\columnwidth]{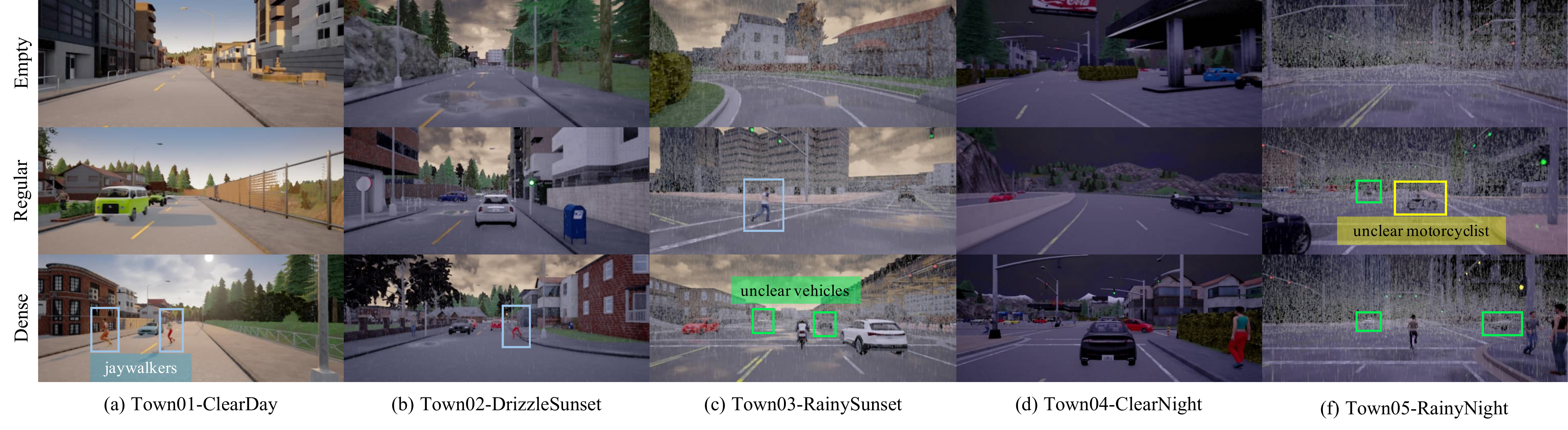}
    \caption{Overview of our dataset: varied maps, weather and illumination conditions with increasing traffic densities (top to bottom). Noticeable road agents are bounded by color boxes. Note that this figure shows only a small part of the environmental setups; please see contexts in Section \ref{dataset_collection} for more details. Columns (a-c) show there can sometimes be jaywalkers running across the roads, for which the ego-vehicle will urgently slow down or completely stop to ensure safety. In addition, it can be seen that in rainy scenarios, especially in \textit{RainyNight}, the surroundings are considerably blurred (e.g., the unclear motorcyclist in the \textit{Regular} setting of column (f)), leading to potential risks for the vision-based driving models\cite{tai2019visual, codevilla2019exploring, codevilla2018end, cai2020vtgnet}.}
    \label{dataset}
    \vspace{-0.3cm}
\end{figure*}

The setups mentioned above can be partially viewed in Fig. \ref{dataset} and more can be viewed in our supplementary videos. These help to generate sufficient interactions between the ego-vehicle and road agents in diverse environments. Based on these setups, we finally collect 360 high-fidelity driving episodes, which last 10.8 hours in total with 389 thousand frames and cover a driving distance of 247 km.

\subsection{Model Architecture}
\subsubsection{Global Planning}
The global planner is separate from the deep networks. It is implemented with the $A^*$ algorithm to plan a high-level coarse route from the start point to the destination based on static town maps. Similar to \cite{pokle2019deep} and \cite{Cai2020HighSpeedAD}, we down-sample the full global route $\bm{G}_f$ to local relevant routes $\bm{G}$ during navigation, which is shown in \eqref{eq:local_route}:
\begin{equation}
    \bm{G} = \left\{\left(x_k,y_k\right)|1\leq k\leq 130 \right\} \subset {\bm{G}_f}.
    \label{eq:local_route}
\end{equation}
Note that the first waypoint $(x_1, y_1)$ in $\bm{G}$ is the closest waypoint in $\bm{G}_f$ to the current location of the vehicle, and the distance of every two adjacent points is 0.4 m. The waypoints are then flattened into a 260-dimensional vector to be processed by dense layers with fully connected ReLU layers. The extracted feature is a higher dimensional vector $\bm{f}_g \in \mathbb{R}^{2048}$.

\subsubsection{Multi-Perception}
With the aim to capture environmental information, the camera records color textures in a 2D image plane, while the lidar captures 3-D spatial locations and the radar records movement information (i.e., speeds of obstacles relative to the ego-vehicle). We combine these sensors together in our network so that the vehicle is able to sense different dimensions of its surroundings. 

\begin{figure}[t]
    \centering
    \includegraphics[width = \columnwidth]{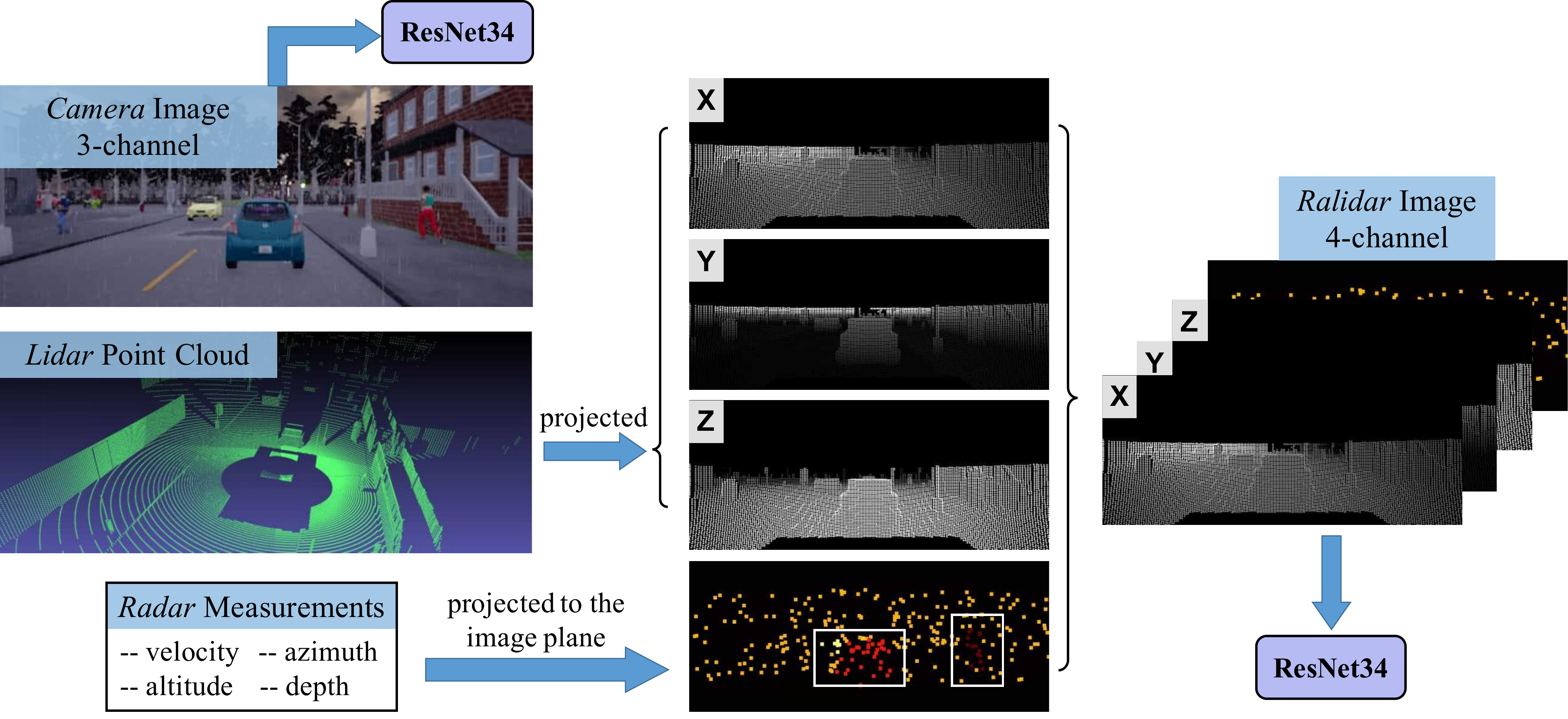}
    \caption{Multimodal data processing. We achieve data alignment by projecting the lidar pointclouds and radar measurements to the image plane by combining them together to form the \textit{ralidar} image. Then, two ResNet34 modules are used to extract features from the camera and ralidar images. Brighter points mean larger values in the projected images. Noticeable road agents in the projected radar image are bounded by white boxes.}
    \label{projection}
    \vspace{-0.3cm}
\end{figure}

Specifically, we project the lidar point clouds and radar data to the image plane with the same width and height as the camera images. We name it the \textit{ralidar} image ($250\times600\times4$), in which the first three channels encode 3-D coordinates and the forth channel encodes relative speeds, as shown in Fig. \ref{projection}. In this way, the multimodal measurements are aligned on the same space and can be uniformly processed with CNNs. In this work, we use ResNet34\cite{he2016resnet} as the backbone to extract environmental features from the camera and ralidar images. The results are feature vectors $\bm{f}_i \in \mathbb{R}^{2048}$ and $\bm{f}_r \in \mathbb{R}^{2048}$.

\subsubsection{End-to-End Action Generation}
In addition to the sensory data and the global route, our network also takes as input the velocity of the ego-vehicle $(v_x, v_y)$ to the dense layers. The extracted feature is a higher dimensional vector $\bm{f}_v \in \mathbb{R}^{2048}$. Then the features $\left[\bm{f}_i,\bm{f}_r,\bm{f}_v,\bm{f}_g\right]$ are handled in two ways: a) we concatenate them into a vector $\bm{f}_c \in \mathbb{R}^{8192}$ for further processing, and b) in the spirit of\cite{pokle2019deep}, we fuse them with an attention mechanism defined in \eqref{eq:attention}, where the coefficients $\bm{a} = [a_i, a_r, a_v, a_g]$ reflect the relative importance of the features in changing environments.
\begin{equation}
    \bm{f}_f = a_i\bm{f}_i + a_r\bm{f}_r + a_v\bm{f}_v + a_g\bm{f}_g.
    \label{eq:attention}
\end{equation}
The coefficients $\bm{a}$ are computed by transforming $\bm{f}_c$ with dense layers and softmax activation. After such feature fusion, a control action $\bm{a}_1$ composed of steering, throttle and brake is generated by projecting $\bm{f}_f$ with fully connected ReLU layers. Inspired by \cite{codevilla2018offline}, we use the L1 loss function for this module as it is better correlated to the online driving performance. 

\begin{table*}[t]
\newcommand{\tabincell}[2]{\begin{tabular}{@{}#1@{}}#2\end{tabular}}
\newcommand{\NA}{---}
        \setlength{\abovecaptionskip}{-1pt}
        \renewcommand{\arraystretch}{1.3}
        \caption{We Evaluate Different Driving Models on Our \textit{DeepTest} Driving Benchmark. $\uparrow$ Means Larger Numbers Are Better, $\downarrow$ Means Smaller Numbers Are Better. The Bold Font Highlights the Best Results in Each Column.}
        \label{tab:evaluation}
        \centering
        \begin{tabular}{l  c c c  c c c  c c c  c c c }
        
        \toprule
        {}&
        \multicolumn{3}{c}{{Training Conditions}}&
        \multicolumn{3}{c}{{New Weather}} & 
        \multicolumn{3}{c}{{New Town}} & 
        \multicolumn{3}{c}{{New Town \& Weather}}\\
        \cmidrule(lr){2-4} \cmidrule(lr){5-7} \cmidrule(lr){8-10} \cmidrule(lr){11-13}
        {Town Name}&
        \multicolumn{3}{c}{\texttt{Town03} (urban)} &
        \multicolumn{3}{c}{\texttt{Town05} (urban)} & 
        \multicolumn{3}{c}{\texttt{Town07} (rural)} & 
        \multicolumn{3}{c}{\texttt{Town06} (urban)}\\
        Traffic Density&
        Empty & Regular & Dense &
        Empty & Regular & Dense &
        Empty & Regular & Dense &
        Empty & Regular & Dense \\
        \hline
        \multicolumn{2}{l}{\textit{Success Rate $\uparrow$ (\%)}}
        \\
        \hline
        CIL\cite{codevilla2018end} & 38 & 16 & 16 & 33 & 11 & 0 & 0 & 0 & 0 & 16 & 11 & 0 \\
        CIL-R & 83 & 55 & 38 & 33 & 22 & 16 & 22 & 11 & 11 & 11 & 11 & 11 \\
        INT\cite{gao2017intention} & 16 & 33 & 11 & 83 & 5 & 5 & 38 & 22 & 5 & 94 & 61 & 16 \\
        PMP (\textit{ours}) & \textbf{100} & \textbf{72} & \textbf{88} & \textbf{100} & \textbf{77} & \textbf{77} & \textbf{100} & \textbf{83} & \textbf{72} & \textbf{100} & \textbf{88} & \textbf{83} \\
        
        \hline
        \multicolumn{2}{l}{\textit{Wrong Lane $\downarrow$ (\%)}}\\
        \hline
        CIL\cite{codevilla2018end} & 66.05 & 45.16 & 50.87 & 57.22 & 64.41 & 46.18 & 35.55 & 36.81 & 40.71 & 44.14 & 52.37 & 52.03 \\
        CIL-R & 26.60 & 25.57 & 19.07 & 26.58 & 36.64 & 41.86 & 8.88 & 7.20 & 3.35 & 42.50 & 50.72 & 51.61 \\
        INT\cite{gao2017intention} & \textbf{0.00} & 0.04 & \textbf{0.01} & \textbf{0.07} & \textbf{0.12} & \textbf{0.15} & \textbf{0.00} & \textbf{0.00} & \textbf{0.00} & \textbf{0.12} & \textbf{0.13} & \textbf{0.28} \\
        PMP (\textit{ours}) & 0.02 & \textbf{0.00} & \textbf{0.01} & 0.40 & 0.48 & 0.50 & 0.04 & \textbf{0.00} & 0.01 & 0.43 & 0.40 & 0.61 \\

        \hline
        \multicolumn{2}{l}{\textit{Overspeed $\downarrow$ (\%)}}\\
        \hline
        CIL\cite{codevilla2018end} & 0.33 & 0.37 & 0.16 & 0.10 & \textbf{0.00} & -- & -- & -- & -- & \textbf{0.04} & \textbf{0.00} & -- \\
        CIL-R & 0.14 & \textbf{0.13} & \textbf{0.08} & \textbf{0.04} & \textbf{0.00} & \textbf{0.00} & 0.33 & \textbf{0.09} & \textbf{0.28} & 0.10 & 0.16 & 1.54 \\
        INT\cite{gao2017intention} & 17.70 & 11.18 & 5.85 & 17.09 & 15.14 & 8.52 & 19.03 & 11.87 & 14.84 & 37.12 & 30.22 & 31.04 \\
        PMP (\textit{ours}) & \textbf{0.11} & 0.22 & 0.12 & 0.14 & \textbf{0.00} & 0.06 & \textbf{0.26} & 0.30 & \textbf{0.28} & 0.40 & 0.28 & \textbf{0.36} \\
        
        \bottomrule
        \end{tabular}
        \vspace{-0.2cm}
\end{table*}

\subsubsection{Probabilistic Motion Planning}
In this module, we aim to learn a full parameterized distribution over possible ego-motions (i.e., velocities and yaw angles) for 3.0 s into the future, as shown in Fig. \ref{network}. We adopt the GMM to represent such a distribution due to its excellent approximation properties. Specifically, the combined feature $\bm{f}_c$ in our work is transformed by dense layers into GMM parameters (i.e., weight, mean and variance) to describe the distribution of future motions. Similar to \cite{amini2019variational} and \cite{huang2019uncertainty}, the negative log-likelihood (NLL) loss function is used for this module.

As mentioned in \cite{wiest2012probabilistic}, the advantage of probabilistic modeling is that we can make a decision by evaluating its statistical properties. In this work, based on the mean values ($\bm{\mu}$) of the planned motion distribution, we further design a PID controller to calculate a control action $\bm{a}_2$ composed of steering, throttle and brake. The target point for this PID controller (assume \textit{k} frames in the future) is set to the point 5 m ahead of the vehicle by calculating the integral with $\bm{\mu}$. Then, the final action $\bm{a}_f$ to control the vehicle is computed by examining the reliability of the motion distribution through its accumulated variance $\bm{\sigma}^2$:
\begin{equation}
    \bm{a}_f = \left(1-\lambda \right)\bm{a}_1 + \lambda\bm{a}_2,\   \lambda = e^{-c_1\cdot max\left(0,\ \sum_i^k\sigma^2 - c_2\right)}.
    \label{eq:action_fusion}
\end{equation}
In this way, higher planning uncertainty leads to smaller $\lambda$, thus the final action will depend more on $\bm{a}_1$. We believe that we can take advantage of both end-to-end control and probabilistic modeling by performing such reliability-aware action fusion.

\section{Experiments and Discussion}

\subsection{Training Setup}

We train the proposed PMP-net on our large-scale driving dataset introduced in Section \ref{dataset_collection}. The full dataset is divided into a training set and a validation set according to the ratio of 7:1, leading to 340K training samples\footnote{Note the test set is not considered because we evaluate our model \textit{online} in Section \ref{sec:evaluation} by making the ego-vehicle directly interact with dynamic environments.}. We use the Adam optimizer with a learning rate of 0.0001, and the batch size is 90. Based on these setups, the model is trained on two Nvidia GeForce RTX 2080 Ti GPUs for about 75 hours, with 234K training steps to achieve convergence. For comparison, we also train and finetune three other baselines on the same training set, which are for visual navigation:
\begin{itemize}
    \item \textbf{CIL}: The conditional imitation learning network introduced in \cite{codevilla2018end}. This maps the camera images and ego-velocities directly to control actions, based on four discrete commands for goal-directed navigation: \textit{follow lane}, \textit{turn left}, \textit{turn right} and \textit{go straight at the intersection}.
    \item \textbf{CIL-R}: We replace the original image processing module of CIL (which is relatively shallow) with ResNet34, to evaluate if deeper models perform better for our task.
    \item \textbf{INT}: The intention-net introduced in \cite{gao2017intention} with the backbone of ResNet34 for fair comparisons. This maps the camera images and global routes to control actions. Note that the original intention-net takes the indoor floor maps rendered with routes for directions. We replace it with the local relevant routes $\bm{G}$ introduced in \eqref{eq:local_route}.
\end{itemize}

\subsection{Evaluation}
\label{sec:evaluation}

\subsubsection{DeepTest Benchmark}
\label{sec:benchmark}

\begin{figure*}[t]
    \centering
    \includegraphics[width = 2\columnwidth]{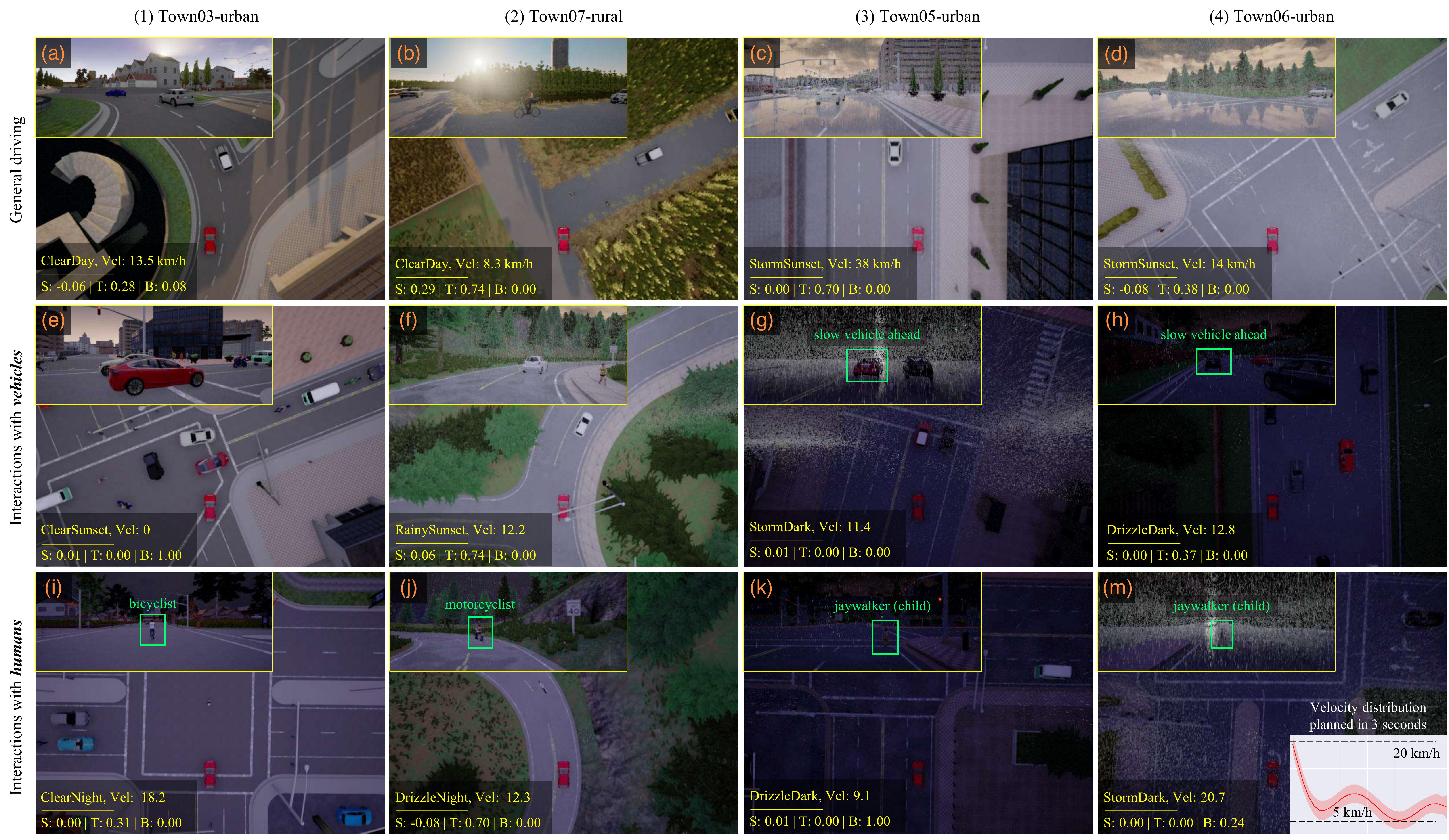}
    \caption{Online evaluation results of PMP-net in our \textit{DeepTest} benchmark. The environment setups, driving velocities and control actions are shown in yellow text. Noticeable road agents (e.g., jaywalkers) are bounded by green boxes. The range of steering is [-1,1], while for throttle and brake the range is [0,1]. The sample driving behaviors are: (c) lane-following, turning at (b,d,e) intersections or (a) roundabouts, (g) lane-changing, (f,h,i,j) vehicle-, bicyclist- or motorcyclist-following, and (k,m) urgently slowing down for jaywalkers. All of these behaviors are performed autonomously and safely by PMP-net in an end-to-end fashion without hand-crafted rules.}
    
    \label{fig:qualitative}
    \vspace{-0.3cm}
\end{figure*}

We evaluate the online driving performance for different models on our proposed \textit{DeepTest} benchmark in CARLA. Compared with the previous benchmarks in \cite{codevilla2019exploring} and \cite{ dosovitskiy2017carla}, \textit{DeepTest} has many more environmental setups, such as more test maps, weather conditions and interactions with road agents. In addition, different to \cite{codevilla2019exploring} and \cite{dosovitskiy2017carla}, we set zero tolerance for collision events, which means that any degree of collisions with static (e.g., trees) or dynamic (e.g., pedestrians) objects leads to a failed episode.

In our benchmark, different methods are tested on four maps. For each map, we set three levels of traffic densities: \textit{empty}, \textit{regular} and \textit{dense}. Therefore, each driving model relates to 12 driving tasks. Note that denser traffic leads to harder driving tasks as it involves more dynamic obstacles on the road. In each task, we further set 18 goal-directed episodes with varied weather conditions. Therefore, to fully evaluate PMP-net and the other three baselines, 864 driving episodes should be conducted. Finally, the evaluation process costs 4 days on our computer and covers a driving distance of 855 km. Compared with the environmental setups in the training set (Section \ref{dataset_collection}), we consider new maps, illuminations and weather in \textit{DeepTest} to test the generalization capability. Specifically, we add an unseen rural map \textit{Town07} and an urban map \textit{Town06}. \textit{Town07} brings new challenges to test the negotiation skills with narrow roads and many non-signalized crossings. In addition, we add four extreme illumination and weather conditions: \textit{ClearDark}, \textit{DrizzleDark}, \textit{StormDark} and \textit{StormSunset}. The new \textit{Dark} and \textit{Storm} (i.e., heavy rain) settings, which are shown in Fig. \ref{fig:qualitative}, bring extra challenges to the drive with limited vision. Similar to \cite{codevilla2018end}, we do not consider traffic lights in this work. For quantification of the driving performance, three metrics are adopted as follows:
\begin{itemize}
    \item \textbf{SR}: success rate. An episode is considered to be successful if the agent reaches a certain goal without any collision within a time limit. Based on this, we calculate the success rate for models in different tasks.

    \item \textbf{WL}: The proportion of the period in a wrong lane to the total driving time.

    \item \textbf{OVSP}: The proportion of the overspeeding period to the total driving time. The speed limit is set to 20 km/h at intersections and 50 km/h elsewhere.
\end{itemize}

\subsubsection{Quantitative Analysis}
We show the results on the \textit{DeepTest} benchmark in Table \ref{tab:evaluation}. In the following, the analyses are given from two perspectives: the \textit{ability} and the \textit{quality} of autonomous driving.

\textbf{Ability:}
Success rate (SR) is used to measure the self-driving ability, which is a crucial concern in this area.

It can be seen that the CIL model presents the worst results, which can not even achieve a successful episode in \texttt{Town07}. In addition, although in \texttt{Town03} we only set new routes with similar environments to the training dataset, CIL still presents low SRs (16\textasciitilde38\%). With the help of a deeper backbone in CIL-R, the performance is improved. For example, the SR in \texttt{Town03-empty} increases from 38\% to 83\%.

By changing the model structure to INT, better generalization performance on certain new environments is achieved, for example, the SR in \texttt{Town06-Regular} increases from 11\% to 61\%. However, INT performs worse than CIL-R in \texttt{Town03} and some other new environments such as \texttt{Town05-Dense}. Generally, INT and CIL-R have similar low-level performances in outdoor driving areas, especially in heavy traffic. This is because they only use visual perception, which often has troubles in tough environments such as \textit{StormDark}. By contrast, PMP-net achieves a much higher SR in all evaluation setups, which indicates a superior generalization capability. In particular, the SR increases to 100\% in all environments for the empty traffic, and to 72\textasciitilde88\% for regular and dense traffic.

\textbf{Quality:}
We use WL and OVSP to evaluate the driving quality of different models. Due to the lack of concrete direction guidance, CIL and CIL-R both have high WL values (3.35\textasciitilde66.05\%). Specifically, they often navigate the vehicle to drive in the correct direction but in the wrong lanes. With the help of the global route information, the models are able to drive more accurately, as we can see by the WL values for INT and PMP, which are all close to 0\%. However, INT tends to control the vehicle to drive at high speeds without slowing down at intersections. This unsafe phenomenon leads to high values of OVSP for INT (5.85\textasciitilde37.12\%). While PMP still performs well on this metric (0.0\textasciitilde0.4\%).

Generally, the remarkable improvements of PMP-net on the benchmark w.r.t. the other three baselines confirm that our proposed model can effectively learn and deploy the driving knowledge in complex dynamic environments.

\subsubsection{Qualitative Analysis}
\label{qualitative}
Fig. \ref{fig:qualitative} shows the qualitative results of PMP-net. When there are no obstacles ahead on straight roads, our model drives relatively fast, at about 40 km/h (Fig. \ref{fig:qualitative}-(c)). When taking turns or following road agents, our model reasonably slows down as a human driver would, as shown in Fig. \ref{fig:qualitative}-(a,b,d,f,i). In addition, we show some results in extreme conditions. In Fig. \ref{fig:qualitative}-(e), the traffic is heavy with many vehicles driving at an intersection. Although the model is directed to turn right, it applies full brake as another vehicle blocks the road ahead. Moreover, in Fig. \ref{fig:qualitative}-(g,h), we set dense traffic on a dark night where slow-moving obstacles are ahead of the ego-vehicle. In these scenes with limited vision, PMP-net is also able to drive safely by reducing the throttle to slow down when changing/following lanes. Furthermore, the most challenging scene is shown in Fig. \ref{fig:qualitative}-(m). In the \textit{StormDark} environment, there is a small child running across the road abruptly without any previous notice. For this scene, it is difficult to raise alarm even for a human driver because the surroundings cannot be seen clearly. Surprisingly, our model slows down timely by applying brake to avoid an accident. Fig. \ref{fig:qualitative}-(k) is another similar scenario. For interpretation, the planned motion distribution of Fig. \ref{fig:qualitative}-(m) is attached, where we can see that the planned speed drops rapidly within a short horizon (\textasciitilde0.5 s) with low variance. We accredit such prominent performance to our multimodal and probabilistic setup. More related driving behaviors are shown in supplementary videos\footnote{\url{https://sites.google.com/view/pmpnet/}}.

\section{Conclusion}
In this paper, to realize autonomous driving in outdoor dynamic environments, we proposed a deep navigation model named PMP-net, which is based on multimodal sensors (a camera, lidar and radar) and probabilistic end-to-end control. We collected a large-scale driving dataset in the CARLA simulator and trained the model with imitation learning. In order to fully evaluate the driving performance, we further proposed a new online benchmark \textit{DeepTest}, of which the environmental complexity has not been previously considered. By setting varied illumination, weather and traffic conditions in different towns, we showed that our model achieves excellent driving and generalization performance in both unseen urban and rural areas with extreme weather and heavy traffic with dynamic objects (e.g., vehicles, bicyclists and jaywalkers). 

To further extend PMP-net for real autonomous vehicles, the \textit{reality gap} should be considered. 1) For discrepancy of sensory input, we can finetune the model with real-world data. The sensor readings of lidar and radar are more consistent than those of a camera with real/simulated deployments, which can help regularize the finetuning process for \textit{domain adaption}. 2) For discrepancy of driving platforms, we can adjust the parameters of the PID controller to adapt to different vehicle properties\cite{cai2020vtgnet}, due to the \textit{modular} design of our network.

\bibliographystyle{IEEEtran}
\bibliography{root.bib}

\begin{figure*}[tb]
    \centering
    \includegraphics[width=2\columnwidth]{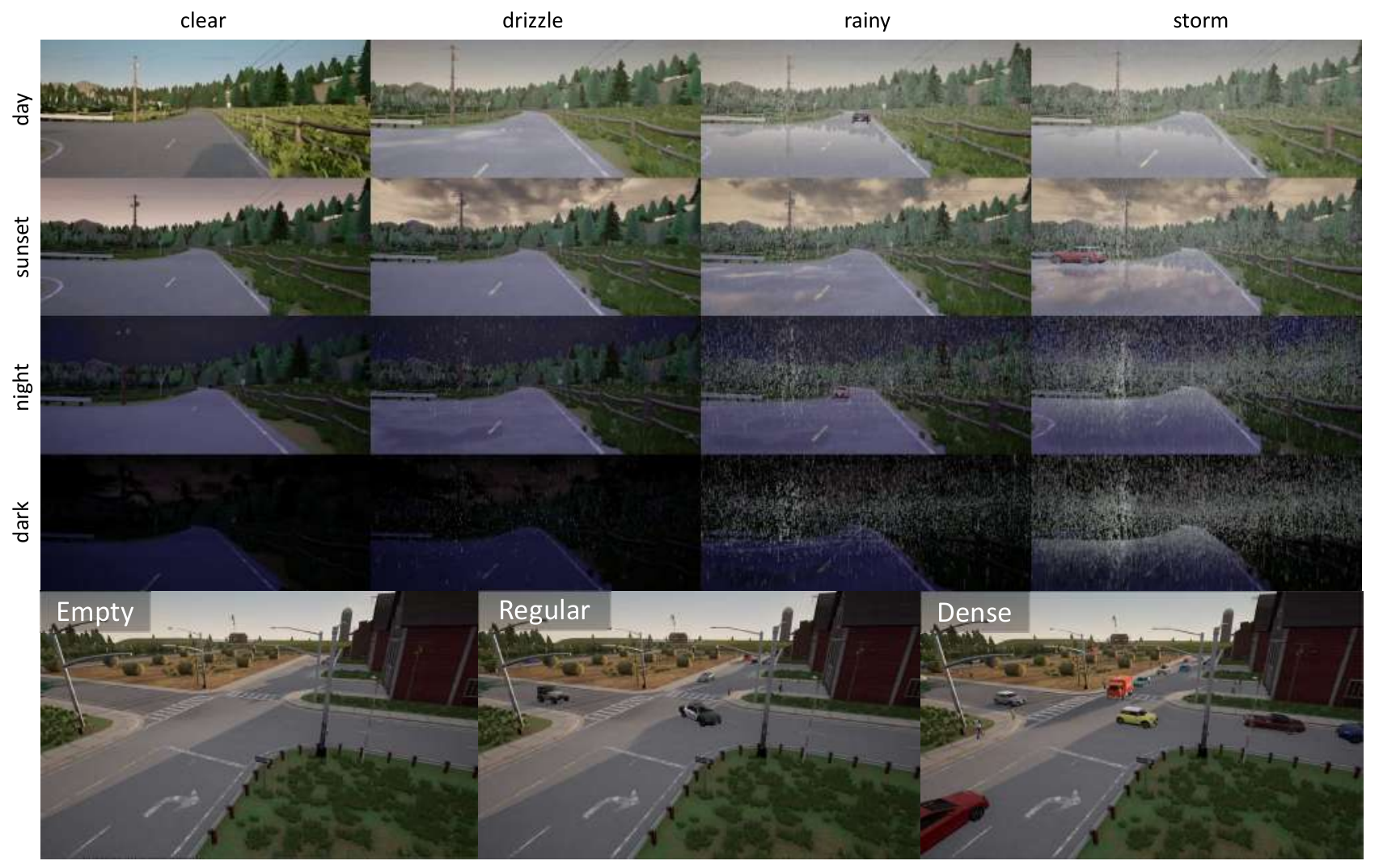}
    \caption{Row1-4: varied illumination (\textit{clear, sunset, night} and \textit{dark}) and weather (\textit{clear, drizzle, rainy} and \textit{storm}) conditions considered in this work; row5: three levels of traffic density in our \textit{DeepTest} benchmark.}
    \label{fig:appendix_fig}
\end{figure*}

\appendix

\subsection{Traffic Densities}
We set three levels of traffic density for our \textit{DeepTest} benchmark: \textit{empty}, \textit{regular} and \textit{dense}. The specific settings are shown in Table. \ref{tab:td} and can be viewed in Fig. \ref{fig:appendix_fig}. 
\begin{table}[t]
\newcommand{\tabincell}[2]{\begin{tabular}{@{}#1@{}}#2\end{tabular}}
\newcommand{\NA}{---}
        \setlength{\abovecaptionskip}{-1pt}
        \renewcommand{\arraystretch}{1.3}
        \caption{Different traffic densities in this work}
        \label{tab:td}
        \centering
        \begin{tabular}{l c c}
        
        \toprule
        Type &Number of pedestrians & Number of vehicles \\
        \midrule
        Empty  & 0 & 0\\
        Regular & 40\textasciitilde75  & 60 \\
        Dense & 60\textasciitilde150 & 80\textasciitilde120\\
        
        \bottomrule
        \end{tabular}
        \vspace{-0.2cm}
\end{table}

\begin{table}[t]
\newcommand{\tabincell}[2]{\begin{tabular}{@{}#1@{}}#2\end{tabular}}
\newcommand{\NA}{---}
        \setlength{\abovecaptionskip}{-1pt}
        \renewcommand{\arraystretch}{1.3}
        \caption{PID parameters used in this work}
        \label{tab:pid}
        \centering
        \begin{tabular}{l c c c}
        
        \toprule
        Type &Proportional (P) & Integral (I) & Derivative (D) \\
        \midrule
        Lateral &0.70 & 0.00 & 0.00\\
        Longitudinal &0.25 & 0.20 & 0.00\\
        
        \bottomrule
        \end{tabular}
        \vspace{-0.2cm}
\end{table}

\subsection{Illumination and Weather Conditions}

This paper involves four illumination conditions (i.e., \textit{daytime}, \textit{sunset}, \textit{night} and \textit{dark}) and four weather conditions (i.e., \textit{clear}, \textit{drizzle}, \textit{rainy} and \textit{storm}). The specific settings can be viewed in Fig. \ref{fig:appendix_fig}.

\subsection{PID Control}
To translate the planned motion distribution into action $\bm{a}_2$, two PID controllers are designed for lateral (steering) and longitudinal (throttle and brake) control. The parameters are shown in Table \ref{tab:pid}.

\end{document}